\documentclass[runningheads]{llncs_cr}
\usepackage[utf8]{inputenc}
\usepackage[english]{babel}
\usepackage[dvipsnames]{xcolor}
\usepackage{amsfonts}
\usepackage{graphicx}
\usepackage{comment}
\usepackage{amssymb} 
\usepackage{color}
\usepackage{booktabs}
\usepackage{caption}
\usepackage{csquotes}
\usepackage{enumitem}
\usepackage{float}
\usepackage{hyperref}
\usepackage{lmodern}
\usepackage{mathtools}
\usepackage{microtype}
\usepackage{multirow}
\usepackage{subcaption}
\usepackage{siunitx}
\usepackage{tabularx}
\usepackage[strings]{underscore}

\DeclareMathOperator*{\argmax}{arg\,max}

\newcommand{\R}{\mathbb{R}}

\newcommand{\E}{\mathbb{E}}

\DeclareCaptionFont{tiny}{\tiny}

\usepackage{xfrac}

\usepackage{xspace}

\newcommand{\squeezeup}{\vspace{-4mm}}

  {\begin{list}{\arabic{enumi}.}%
     {\topsep=0in\itemsep=0in\parsep=0in\partopsep=0in\usecounter{enumi}}%
   }{\end{list}}

\begin{document}
\pagestyle{headings}
\mainmatter

\title{Vax-a-Net: Training-time Defence  Against Adversarial Patch Attacks}
\titlerunning{Vax-a-Net: Training-time Defence  Against Adversarial Patch Attacks}
\authorrunning{Proc. ACCV 2020.  T. Gittings, et al.}

\author{T. Gittings\inst{1}, S. Schneider\inst{2} and J. Collomosse\inst{1}}

\institute{Centre for Vision Speech and Signal Processing (CVSSP),\\University of Surrey, Guildford, UK. \and
Surrey Centre for Cyber Security (SCCS),\\University of Surrey, Guildford, UK.  \\ \email{\{t.gittings,s.schneider,j.collomosse\}@surrey.ac.uk}}

\maketitle

\begin{abstract}
We present Vax-a-Net; a technique for immunizing convolutional neural networks (CNNs) against adversarial patch attacks (APAs).  APAs insert visually overt, local regions (patches) into an image to induce misclassification.  We introduce a conditional Generative Adversarial Network (GAN) architecture that simultaneously learns to synthesise patches for use in APAs, whilst exploiting those attacks to adapt a pre-trained target CNN to reduce its susceptibility to them. This approach enables resilience against APAs to be conferred to pre-trained models, which would be impractical with conventional adversarial training due to the slow convergence of APA methods. We demonstrate transferability of this protection to defend against existing APAs, and show its efficacy across several contemporary CNN architectures.
\end{abstract}

\section{Introduction}
\begin{figure}[t]
	\centering{\includegraphics[width=\textwidth]
		{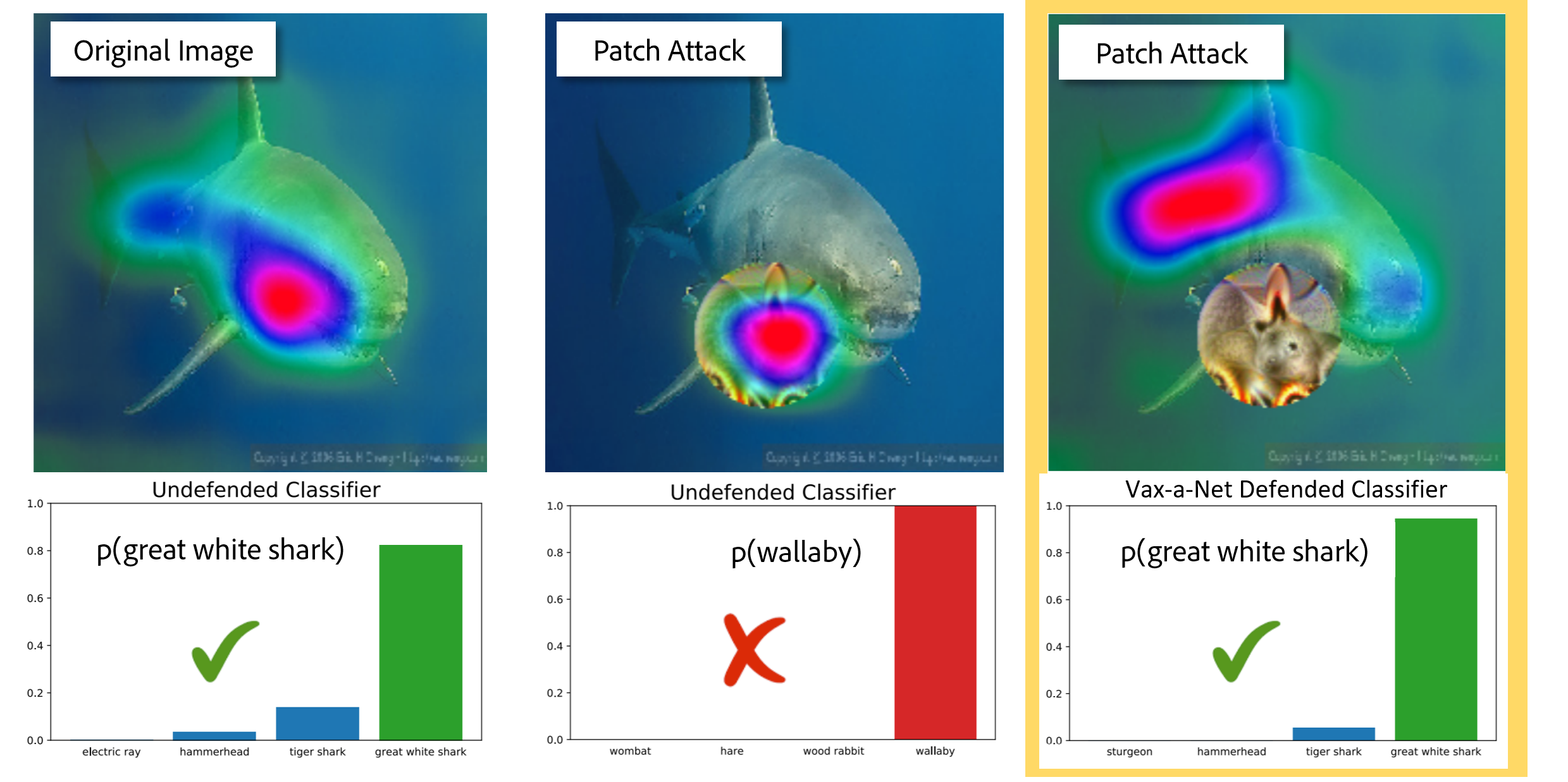}}
	\caption{\label{fig:headline} Vax-a-Net vaccinates pre-trained  CNNs against adversarial patch attacks (APAs); small image regions crafted to induce image misclassification. A shark is correctly classified by a VGG CNN (left), but fooled by an APA \cite{Brown2017} (middle). Vax-a-Net applies defensive training to improve the CNN's resilience to the APA (right). Visualizations show CNN attention (via Grad-CAM \cite{gradcam}). }
	\squeezeup
\end{figure}
Convolutional neural networks (CNNs) are known to be vulnerable to adversarial examples: minor changes made to an image that significantly affect the classification outcome \cite{Szegedy2013,Goodfellow2014}.  Adversarial examples may be generated by pixel-level perturbation of the image, introducing covert yet fragile changes that induce misclassification \cite{Goodfellow2014,DeepFool,CarliniWagner,Gittings2019}. More recently, adversarial patches or ‘stickers’ have been proposed \cite{Brown2017,eykholt2018robust,Gittings2019}, creating overt changes within local image regions that exhibit robustness to affine transformation, and even to printing.  Despite the increasing viability of such `adversarial patch attacks’ (APAs) to confound CNNs in the wild, there has been little work exploring defences against them.

The core contribution of this paper is a new method to defend CNNs against image misclassification due to APAs.  Existing defences typically seek to detect and remove patches in a pre-processing step prior to inference; \emph{e.g.}\ exploiting the high visual salience of such patches.  Yet the manipulation or removal of salient content often degrades model performance (c.f. Table~\ref{tbl:control}). To avoid these problems we propose adapting the method of adversarial training to the realm of APAs. We leverage the idea of generative adversarial networks (GANs) \cite{goodfellow2014generative} to simultaneously synthesise effective adversarial patches to attack a target CNN model, whilst fine-tuning that target model to enhance its resilience against such attacks.  Existing APA methods synthesise a patch via optimizations that take several minutes to converge \cite{Brown2017,Gittings2019}. In order to incorporate patch synthesis into the training loop, patch generation is run via inference pass on the Generator which takes less than one second.  Furthermore, patch generation is also class-conditional; a single trained generator can create patches of many classes. Moreover, we demonstrate that the protection afforded to the model transfers to also defend against existing APA techniques \cite{Brown2017,Gittings2019}.

We show for the first time  that  adversarial training may be leveraged to adapt a pre-trained CNN model's weights to afford it protection against state of the art APAs.  We demonstrate this for both untargeted attacks (seeking misclassification) and targeted attacks (seeking misclassification to a specific class) over several contemporary CNN architectures. We demonstrate that a CNN may be `vaccinated' against two state of the art APA techniques \cite{Brown2017,Gittings2019} despite neither being invoked in that process. Immunising a CNN model against APA via further training, contrasts with  existing APA defences  that  filter images to mitigate patches at inference time. We show our method better preserves classification accuracy, and has a higher defence success rate than inference-time defences \cite{Naseer2019,Hayes2018}.

The adoption of CNNs within safety-critical autonomous systems opens a new facet of
cyber-security, aimed on one hand to train networks resilient to adversarial attacks, and on the other to evaluate resilience by developing new attacks.  This paper makes explicit that connection through adversarial training to immunise CNNs against this emerging  attack vector.

\section{Related Work}
Szegedy \emph{et al.}\ introduced adversarial  attacks through minor perturbations of pixels  \cite{Szegedy2013} to induce CNN image misclassification. Goodfellow \emph{et al.}\ later introduced the fast gradient sign method (FGSM, \cite{Goodfellow2014}) to induce such perturbations quickly in a single step, exploiting linearity of this effect in input space. These methods require access to the target model in order to backpropagate gradients to update pixels, inducing high frequency noise that is fragile to  resampling. Later work improved robustness to affine transformation \cite{DeepFool}, whilst  minimising perceptibility of  the perturbations \cite{CarliniWagner}. Gittings \emph{et al.}\ \cite{Gittings2019} improved robustness using Deep Image Prior \cite{Ulyanov2018} to regularise perturbations to the manifold of natural images. Nevertheless current attacks remain susceptible to minor scaling or rotation. Other work made use of generative architectures to produce more effective attacks \cite{baluja2018learning,song2018constructing,bai2020ai,xiao2018generating}.

 {\bf Adversarial Patch Attacks.} Brown \emph{et al.}\ demonstrated that adversarial patches could be used to fool classifiers; they restricted the perturbation to a small region of the image and explicitly optimised for robustness to affine transformations \cite{Brown2017}. Both Brown \emph{et al.}, and later Gittings \emph{et al.}\ \cite{Gittings2019} backpropagate through the target model to generate `stickers' that can be placed anywhere within the image to create a successful attack.  This optimization process can take several minutes for one single patch. Karmon \emph{et al.}\ showed in LaVAN that the patches can be much smaller if robustness to  affine transformation is not required \cite{Karmon2018} but require pixel-perfect positioning of the patch which is impractical for real APAs. In the complementary area of object detection (rather than image classification, addressed this paper) Liu \emph{et al.}\ disabled an object detector using a small patch in one corner of the frame \cite{Liu2018}. Eykholt \emph{et al.}\ applied adversarial patches to traffic signs, explicitly optimising for printability \cite{eykholt2018robust}. Chen \emph{et al.}\ performed a similar attack on an object detector with Stop signs \cite{Chen2018}. Thys \emph{et al.}\ attacked a person detector using a printable patch \cite{Thys2019}.

 {\bf Defences at Training Time.} Whilst introducing adversarial examples, Szegedy \emph{et al.}\ also proposed {\em adversarial training} to defend against them \cite{Szegedy2013}.  Adversarial training  is a form of data augmentation that introduces adversarial examples during the training process in order to promote robustness. This method was impractical when first proposed due to the slow speed of producing adversarial examples making it infeasible to do so during training, but this was resolved by Goodfellow \emph{et al.}'s FGSM \cite{Goodfellow2014}, and later others with  more general fast gradient methods \cite{Lyu2015,Shaham2018}.
Kurakin \emph{et al.}\ applied adversarial training to the ImageNet dataset for the first time \cite{Kurakin2016}. Jang \emph{et al.}\ make use of a recursive attack generator for more effective adversarial training on MNIST and CIFAR-10 \cite{Jang2019}. Papernot \emph{et al.}\ applied the idea of distilling the knowledge of one neural network onto another in a way that masks the gradients at test time and prevents an attacker from being able to use backpropagation \cite{Papernot2016}. All the above only train or fine-tune models to defend against adversarial image examples, rather than defending against localised patch attacks i.e. APAs as in our work.

{\bf Defences at Inference Time.} Meng and Chen observed that by approximating the manifold of natural images it is possible to remove  perturbations within an adversarial image as a pre-process at inference time. By projecting the full image onto this manifold \cite{Meng2017a}; they approximated the input image using an autoencoder. Samangouei \emph{et al.}, and separately Jalal  \emph{et al.}, use a GAN in place of an autoencoder \cite{samangouei2018defensegan,Ilyas2017} to similarly remove adversarial perturbations.

Naseer \emph{et al.}\ \cite{Naseer2019} have created one of the few defences against localised perturbations \emph{i.e.}\ APAs. They observe that adversarial patches are regions of the image with especially high gradient (this is likely how they draw attention  over other areas of the image). By applying local gradient smoothing (LGS) -- conceptually the opposite of a bilateral/edge-preserving blur -- patches are neutralised but at the cost of lowering the classification accuracy on clean images, since classifiers rely upon structural edge detail as a recognition cue.
Hayes \cite{Hayes2018} created a different method to defend against localised adversarial attacks. The defence is split into two stages: detection and removal. To detect the patch they create a saliency map using guided backpropagation and assume that a collection of localised salient features implies that there is a patch. To remove the patch, an image in-painting algorithm  \cite{telea2004image} is applied to the masked region cleaned up via some morphological filtering. 

Rather than attempt to detect and erase adversarial patches, Vax-a-net takes a generative adversarial approach to simultaneously create attack patches and fine-tune the model to `vaccinate' it against APAs.

\section{Method}
\begin{figure}[t!]
	\centering{\includegraphics[width=\textwidth]
		{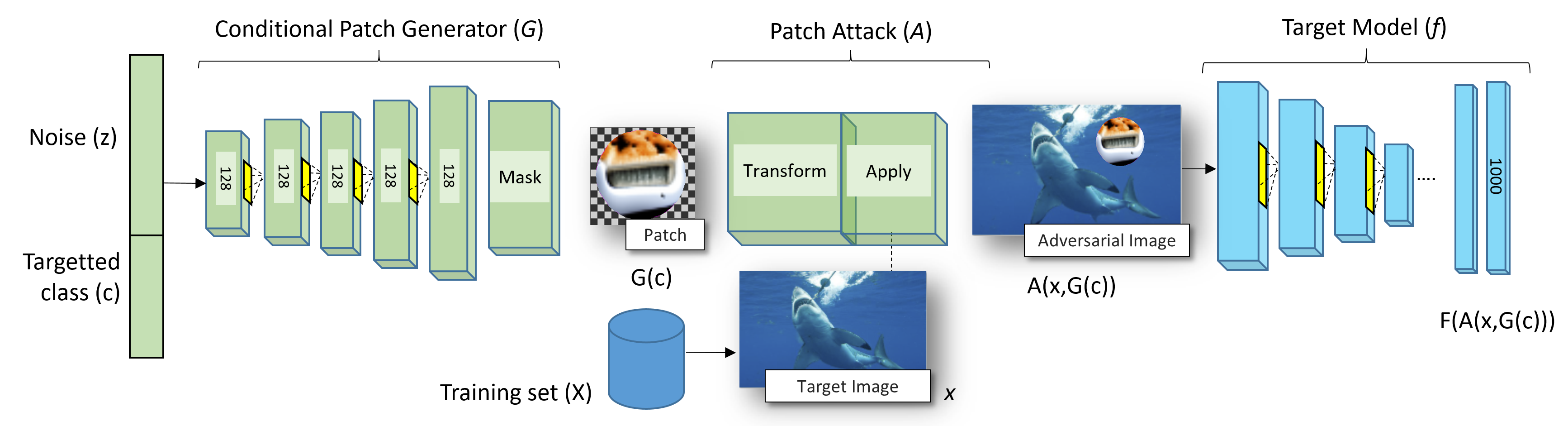}}
	\caption{\label{fig:patch_gan_arch} Proposed architecture for using adversarial training to robustify a model \(f\) against adversarial patch attacks. The conditional patch generator \(G\) can synthesise adversarial patches for \(f\) attacking multiple classes. We alternately train \(G\) and \(f\) to promote the resilience of the model against APAs \cite{Brown2017,Gittings2019}.}
	\squeezeup
\end{figure}

\begin{figure}[t]
	\centering{\includegraphics[width=\textwidth]
		{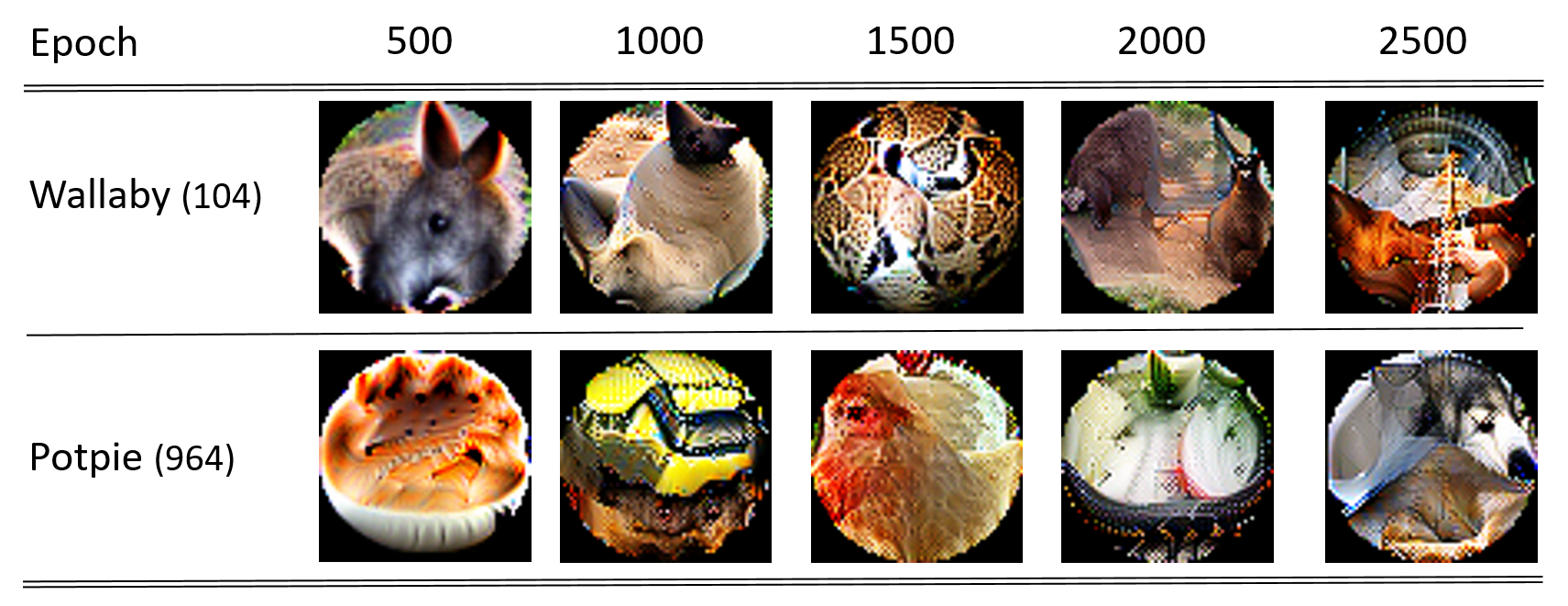}}
	\caption{\label{fig:patchfig} Representative patches sampled from our generator at training epochs 500-2500 for two attack classes.  Patches were generated to defend a VGG-19 model trained on ImageNet.}
	\squeezeup
\end{figure}

Consider a CNN classifier \(f:\R^m\to\R^m\) pre-trained to map a source image $x$  to vector of probabilities $f(x)$, encoding the chance  of the image containing each of a set of classes $c \in \mathcal{Y}$. Adversarial image attacks introduce a perturbation \(r\in\R^m\) to that source image such that \(\argmax_i(f_i(x+r))\neq \argmax_i(f_i(x))\).  We say such  attacks are {\em untargeted}; seeking only to induce misclassification.  If our aim is to introduce a perturbation  $r$ such that \(\argmax_i(f_i(x+r))=c\) we say the attack is {\em targeted} to a specific class ($i$).

Most adversarial images $x+r$ are covert attacks; typically a barely perceptable $r$, distributed across the whole image, is sought.  By contrast, adversarial patch attacks (APAs) have been introduced as overt attacks, in which an adversarial patch (`sticker') is synthesised and composited into a region of an image in order to induce misclassification.  We define a region of interest (ROI) via binary mask $M \in [0,1]$.  In this case we seek perturbation $r$, which can be large, to create a composite image
\begin{equation}
\hat{x} = M \odot r + (1-M) \odot x 
\end{equation}
where $\odot$ is element-wise multiplication. A single adversarial patch capable of attacking multiple images can be created by sampling $x$ in mini-batches from a set of training images (versus learning $r$ over a single image, as is typical for whole image case), as we now explain.

\subsection{Conditional Patch Generation}
Our aim is to defend a pre-trained CNN classifier model against adversarial patch attacks exclusively through modifications in the training process. Although this has been achieved with good success for adversarial image examples, the process of adversarial training used in that case does not apply straightforwardly to the case of adversarial patches. Existing methods of adversarial training require patches to be synthesised at each step of the training process, which is impractical as existing APA methods can take several minutes to synthesise patches. To mitigate this, we adapt the idea of a conditional Deep Convolutional Generative Adversarial Network (DC-GAN) \cite{radford2015unsupervised}, to synthesise effective adversarial patches while simultaneously training the model to defend against those patches.

Fig.\ \ref{fig:patch_gan_arch} illustrates the Vax-a-Net architecture; a conditional patch generator \(G\) is used to synthesise patches which are then applied via a differentiable affine transformation and compositing operation to a training image. The training image is then classifed via the target CNN $f$ which we wish to defend; this model plays the role of discriminator in the GAN.

Our conditional patch generator \(G\) takes an input of a noise vector \(z\), accompanied by a one-hot vector encoding the class $c$ that the attack is  targeting, and produces an adversarial patch of size \(64\times64\). It consists of five up-convolutional layers with filter size \(4\times4\). The number of output channels for the hidden layers are 1024, 512, 256, 128 respectively. The first layer has a stride of 1 and no padding, the remainder have a stride of 2 and 1 pixel of zero-padding. We use batch normalisation after each  layer, and leaky-ReLu activation. Our proposed loss function for the generator is
\begin{equation}\label{eq:gen_loss}
L_G = \E_{c,z,x,t,l} J(f(A(G(z,c), x, l, t)),c),
\end{equation}
where \(A\) is the patch application operator, which we will define and explain further in Sec. \ref{sec:application}, and \(J\) is the cross-entropy loss between the output of \(f\) and the target class.

\begin{figure}[htbp]
	\centering{\includegraphics[width=0.85\linewidth]
		{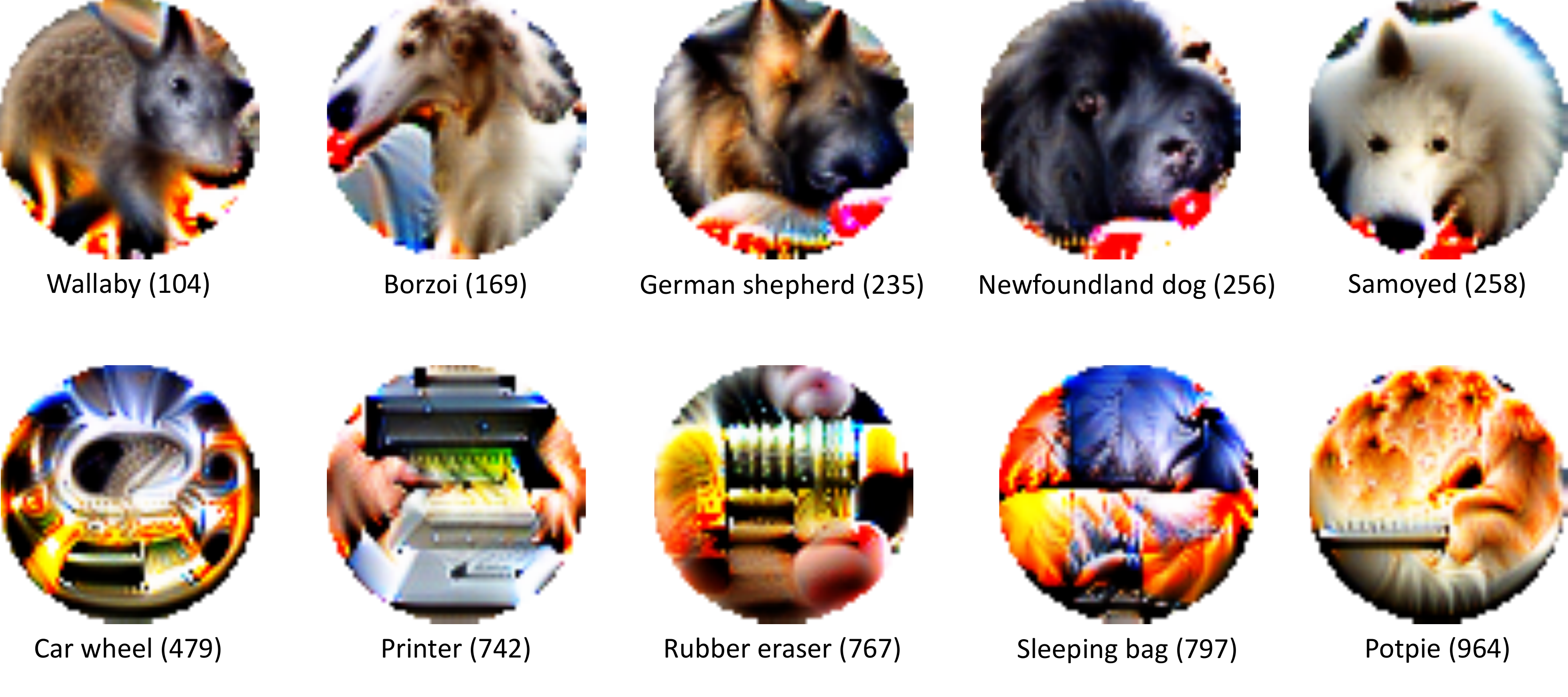}}
	\caption{\label{fig:conditional_patches} Patches sampled from our conditional generator $G$ to attack an undefended VGG-19 model. }
	\squeezeup
\end{figure}
 In our work we explore \(G\) capable of producing effective patches for 1-50 different ImageNet classes (c.f. subsec.\ \ref{sec:classexp}). Fig.\ \ref{fig:conditional_patches} shows the patches that a conditional generator for 10 classes can produce after 500 epochs of training without training the discriminator, i.e.\ these are patches effective at attacking the undefended network. Fig.~\ref{fig:patchfig} shows how patch content evolves as training proceeds beyond the initial training, taking into account  the  discriminator.  The patches resemble abstract versions of the object they are attacking, but with striking colour to attract attention away from other objects.

\subsection{Patch Application and Target Model}
\label{sec:application}
The output of our generator $G(z,c)$ is an image of size \(64\times64\), which we must turn into a patch and apply to the image. First we apply a circular mask to create a round patch (after \cite{Gittings2019},\cite{Brown2017}). Next we apply the patch \(p\) to the image \(x\) at location \(l\) and with an affine transformation \(t\). We denote the output of this operation as \(A(p, x, l, t)\). We use an expectation over transformation to ensure the patch works in any location and with any affine transformation applied. In our training, we enable random rotation of up to \(\sfrac{\pi}{4}\), scaling to between \(1\%\) and \(25\%\) of the image, and translation to any location on the image.

The training process consists of two stages. Initially the discriminator (classifier) is frozen, and we train our generator to produce effective adversarial patches. We then alternate between training the generator and discriminator for each batch, in the usual manner for training a GAN. 

The loss function for \(G\) was defined in Equation \ref{eq:gen_loss}. Our loss function for \(f\) is
\begin{equation}
L_f=\E_{c,z,x,w,t,l} (J(f(A(G(z,c), x, l, t)),y) + J(f(x), y) + \lambda J(f(w), c)),
\end{equation}
where \(w\) are images of class \(c\). Recall that \(J(f(x), y)\) is the cross-entropy loss between the output of CNN \(f\) applied to classify the image \(x\) and the ground truth class \(y\). In practice to approximate the expectation we sample $x$ in mini-batches from a set of training images, and for each image we randomly pick \(c\neq y\) from our set of attack classes (Sec.~\ref{sec:datasets}), \(l\), \(t\) from fixed distributions \(\mathcal{L}\), \(\mathcal{T}\), and \(z\) from a standard normal distribution. The first term of the loss ensures that the model correctly classifies images with patches, the second ensures that the model continues to correctly classify images without patches, and the third is to ensure that it continues to correctly classify images of class \(c\). We empirically selected the weight \(\lambda\) of the third term to have a value of 2.

\subsection{Training Methodology}
The architecture of our generator is close to standard for a GAN, and in place of the discriminator we have a CNN classifier which we intend to robustify. Instead of using the discriminator as a tool to enable the generator to learn how to sample from some underlying distribution from which the training data are drawn (e.g. the distribution of natural images),  we are using a similar architecture to perform a different task. The main difference stems from  our final goal; to end up with a discriminator that is not fooled by any patches (hence a generator with a low success rate), which is the opposite of a regular GAN. Another difference is that our discriminator is a classifier for many (here, 1000 ImageNet classes) not a binary classifier for real/fake, again meaning that the generator will never be able to achieve its goal since the goalposts constantly move \emph{i.e.}\ there is no underlying static distribution that it will approximate.

We pre-train the generator for 500 epochs  before alternating the training of both for each batch. For the generator we use an Adam optimiser with learning rate 0.001 and for the discriminator, Adam  with learning rate of \num{2e-7}.

\begin{figure}[t!]
	\centering %
\begin{subfigure}[b]{0.45\textwidth}
	\includegraphics[width=\textwidth]{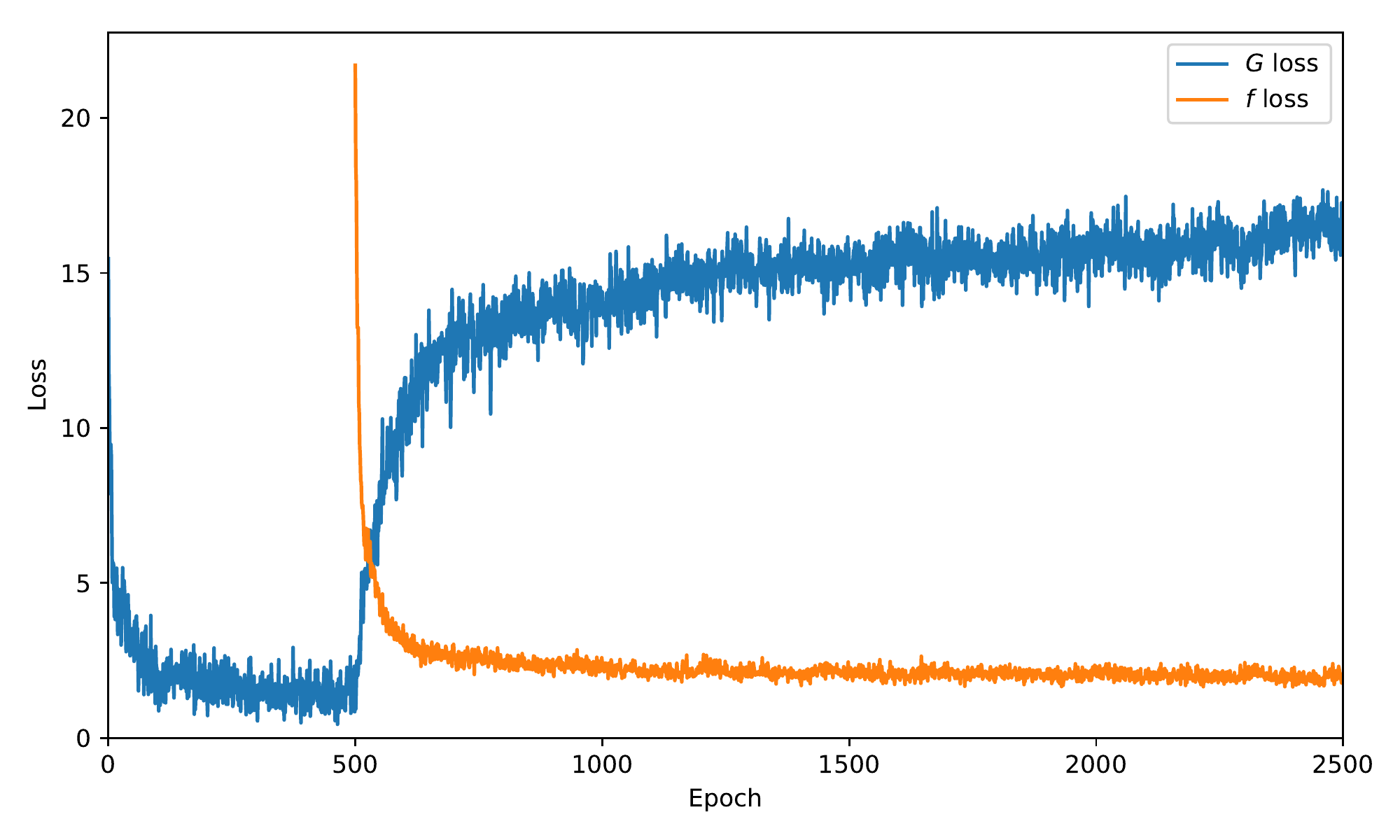}
	\caption{Training Losses}
	\label{fig:train_loss}
\end{subfigure}
\begin{subfigure}[b]{0.45\textwidth}
	\includegraphics[width=\textwidth]{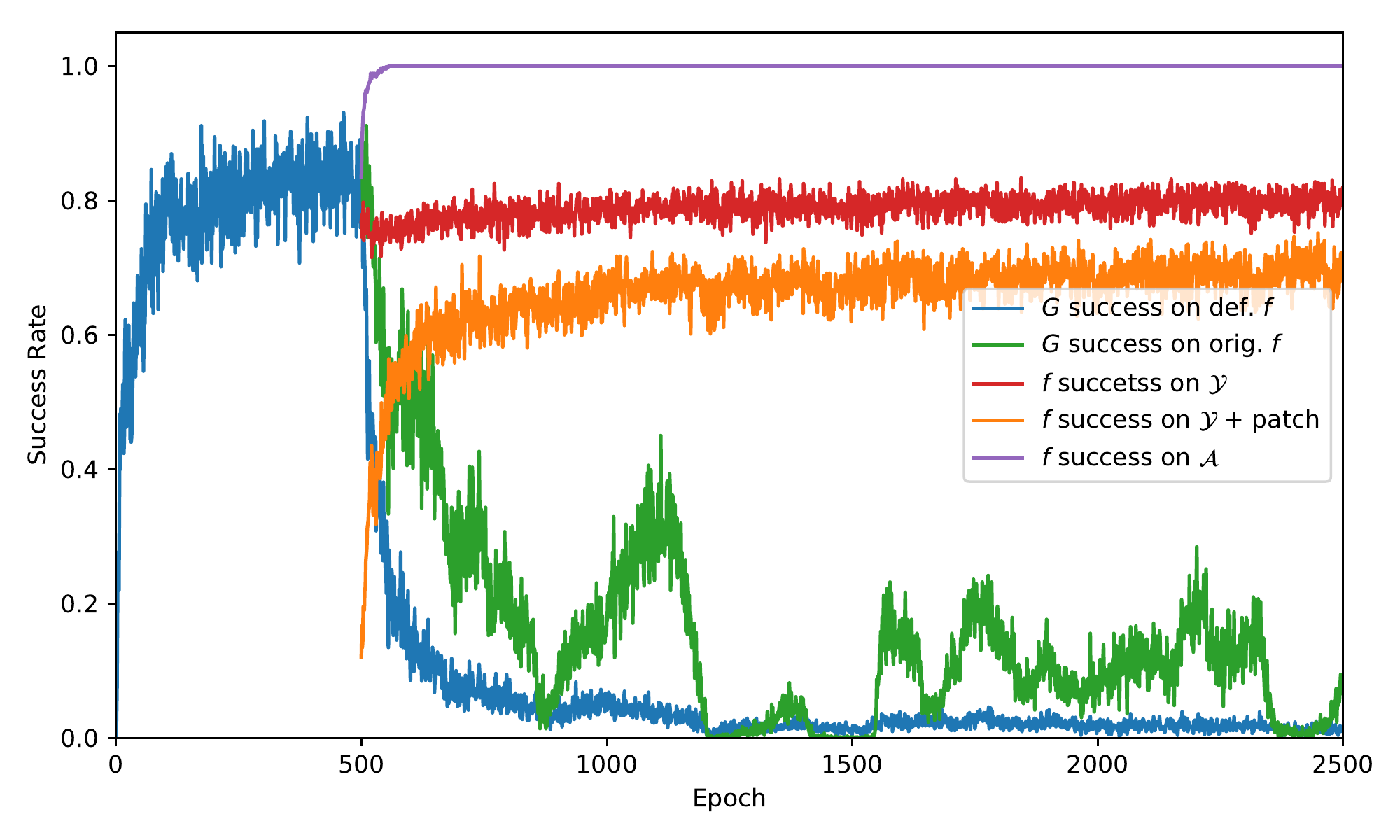}
	\caption{Training Success Rates}
	\label{fig:train_succ}
\end{subfigure}
	\caption{\label{fig:training} Training losses and success rates for our VaN defence. (a) shows the losses for \(G\) and \(f\). Recall that for the first 500 epochs \(f\) is not trained, hence why its line is missing. (b) shows the training success rates of patches from \(G\) applied to the current \(f\) (blue), as well as the original \(f\) (green). It also shows the success of \(f\) at classifying images from \(\mathcal{Y}\), with (orange) and without (red) patches, and also \(\mathcal{A}\) (purple) see Sec. \ref{sec:experiments}}
	
	\squeezeup
\end{figure}

Fig.\ \ref{fig:training} shows both the losses and the success rates on the training data for both \(G\) and \(f\). We observe that during the 500 epoch pre-training phase for \(G\) its loss $L_G$ becomes close to zero and its attack success rate climbs to $\sim80\%$, showing that we can produce effective adversarial patches with our conditional generator. Once the discriminator is updated, it quickly learns not to be fooled by the patches, so the success rates for \(f\) increase while those for \(G\) decrease. The success rate of patches produced by \(G\) when applied to the original model is quite erratic, but declines over time. This confirms that \(f\) is diverging from its original state, and  that the set of patches effective at fooling it diverges from those that original fooled the undefended model.

\section{Experiments and Discussion}
\label{sec:experiments}
\begin{table}[t!]
\centering
\small
\caption{\label{tbl:control}Control: Accuracy of models over the set of test images without attacks $\hat{\mathcal{I}}$, reported for all ImageNet classes ($\mathcal{Y}$) and the subset of these classes used to form patches for APA ($\mathcal{A}$). Reported as top-1 accuracy for the undefended model, and the model defended by our method (D-VaN) or baselines. }
\begin{tabular}{l|ccc|ccc}
\toprule 
\multirow{2}{*}{Method} & \multicolumn{3}{c|}{All classes $\mathcal{Y}$} &  \multicolumn{3}{c}{Attack classes $\mathcal{A}$ \cite{Gittings2019}} \\ 
& VGG & Inception & IRN-v2 & VGG & Inception & IRN-v2 \\
    \midrule
Undefended & 0.692 & 0.770 & 0.788 & 0.616 & 0.704 & 0.772 \\
D-VaN/Ours & \textbf{0.725} & \textbf{0.772} & \textbf{0.803} & \textbf{0.908} & \textbf{0.868} & \textbf{0.884} \\
D-WM & 0.492 & - & 0.523 & 0.396 & - & 0.476 \\
D-LGS & 0.476 & 0.688 & 0.708 & 0.492 & 0.660 & 0.692 \\
\bottomrule
\end{tabular}

\end{table}

\begin{table}[t!]
\centering
\small
\caption{Success rate of defences against adversarial patch attacks covering \(10\%\) or \(25\%\) of the image. We report figures for our Vax-a-Net defence (D-VaN) as well as baseline defences and undefended models. The defence success rate is the proportion of images  classified correctly despite the application of APA (higher is better).}
\begin{tabular}{l|c|cccc}
\toprule 
    \multirow{2}{*}{Architecture} & \multirow{2}{*}{Defence} & \multicolumn{2}{c}{A-ADS \cite{Brown2017}} & \multicolumn{2}{c}{A-DIP \cite{Gittings2019}} \\
    & & 10\% & 25 \% & 10\% & 25\% \\
    \midrule
    \multirow{5}{*}{VGG}& Undefended & 0.041 & 0.006 & 0.016 & 0.001 \\
& D-VaN(D) & 0.410 & 0.147 & 0.422 & 0.154 \\
& D-Van(U) & \textbf{0.642} & \textbf{0.495} & \textbf{0.643} & \textbf{0.483} \\
& D-WM & 0.232 & 0.136 & 0.212 & 0.101 \\
& D-LGS & 0.120 & 0.020 & 0.115 & 0.008 \\
    \midrule
\multirow{4}{*}{Inception}& Undefended & 0.068 & 0.014 & 0.082 & 0.028 \\
& D-VaN(D) & 0.513 & 0.235 & 0.537 & 0.303 \\
& D-VaN(U) & \textbf{0.684} & \textbf{0.541} & \textbf{0.689} & \textbf{0.542} \\
& D-LGS & 0.237 & 0.069 & 0.201 & 0.066 \\
    \midrule
\multirow{5}{*}{IRN-v2}& Undefended & 0.093 & 0.023 & 0.087 & 0.035 \\
& D-VaN(D) & 0.607 & 0.350 & 0.546 & 0.299 \\
& D-VaN(U) & \textbf{0.750} & \textbf{0.642} & \textbf{0.746} & \textbf{0.628} \\
& D-WM & 0.455 & 0.365 & 0.438 & 0.347 \\
& D-LGS & 0.252 & 0.072 & 0.218 & 0.060 \\
\bottomrule
\end{tabular}
\label{tbl:defence_success}
\end{table}
\begin{table}[t!]
\centering
\small
\caption{Success rate of attacks against our models defended by Vax-a-Net, as well as models defended with the baselines, and undefended models. The attack success rate is the proportion of images classified as the adversarial target class when APA is applied (lower is better).}
\begin{tabular}{l|c|cccc}
\toprule 
    \multirow{2}{*}{Architecture} & \multirow{2}{*}{Defence} & \multicolumn{2}{c}{A-ADS \cite{Brown2017}} & \multicolumn{2}{c}{A-DIP \cite{Gittings2019}} \\
    & & 10\% & 25 \% & 10\% & 25\% \\
    \midrule
    \multirow{5}{*}{VGG}& Undefended & 0.910 & 0.990 & 0.962 & 0.999 \\
& D-VaN(D) & 0.053 & 0.385 & 0.075 & 0.353 \\
& D-VaN(U) & \textbf{0.012} & \textbf{0.031} & \textbf{0.010} & \textbf{0.035} \\
& D-WM & 0.516 & 0.544 & 0.553 & 0.661 \\
& D-LGS & 0.553 & 0.903 & 0.577 & 0.952 \\
    \midrule
\multirow{4}{*}{Inception}& Undefended & 0.880 & 0.979 & 0.871 & 0.953 \\
& D-VaN(D) & 0.047 & 0.332 & 0.027 & 0.214 \\
& D-VaN(U) & \textbf{0.016} & \textbf{0.046} & \textbf{0.008} & \textbf{0.017} \\
& D-LGS & 0.557 & 0.765 & 0.645 & 0.833 \\
    \midrule
\multirow{5}{*}{IRN-v2}& Undefended & 0.884 & 0.949 & 0.881 & 0.923 \\
& D-VaN(D) & \textbf{0.004} & 0.180 & 0.018 & 0.238 \\
& D-VaN(U) & 0.005 & \textbf{0.016} & \textbf{0.005} & \textbf{0.020} \\
& D-WM & 0.304 & 0.284 & 0.314 & 0.267 \\
& D-LGS & 0.587 & 0.825 & 0.679 & 0.875 \\
\bottomrule
\end{tabular}
\label{tbl:attack_success}
\end{table}

We evaluate our proposed Vax-a-Net (VaN) method for defending against adversarial patch attacks (APAs) on image classification models trained using three popular network architectures; VGG-19 \cite{simonyan2014very}, Inception-v3 \cite{szegedy2016rethinking}, and Inception-ResNet-v2 (IRN-v2) \cite{szegedy2017inception}.  

{\bf Baselines.} We compare the efficacy of our Vax-a-Net defence ({\bf D-VaN}) against 2 baseline APA defences: the local gradient smoothing ({\bf D-LGS}) method of Naseer \emph{et al.}\ \cite{Naseer2019} and the watermark removal method ({\bf D-WM}) of Hayes \cite{Hayes2018}. We test the effectiveness of our defence and the baseline defences against 2 baseline patch attacks; the adversarial stickers ({\bf A-ADS}) method of Brown \emph{et al.}\ \cite{Brown2017}, and the deep image prior based ({\bf A-DIP}) method of Gittings \emph{et al.}\ \cite{Gittings2019}. For all attacks we used public open source implementations, but for  defences due to absence  of author code we use our own implementations in the open-source PyTorch library \cite{pytorch}. Due to the architecture of the pre-trained network available in PyTorch and the nature of the defence we were unable to implement D-WM on the Inception-v3 model, and results for this model were not originally reported.
\label{sec:datasets}
{\bf Datasets.} We evaluate over the ImageNet \cite{Deng2009} dataset containing 1k object classes $\mathcal{Y}$, using the published training  (1.2M images) and test (50k images; 50 per class) partitions. 
For each of the architectures tested we use a model pre-trained on ImageNet, distributed with PyTorch. We refer to these as {\bf undefended models}. Our proposed defence (D-VaN) involves further training of undefended models using the same training set. The test set comprises 50k images upon which attacks are mounted, each by inserting one adversarial patch. Let this unaltered test set be $\hat{\mathcal{I}}$.  The patch is crafted to encourage an image containing object of ground truth class $y \in \mathcal{Y}$ to be  misclassified a single target class $c \in \mathcal{A}$; we use the subset of 10 attack classes $\mathcal{A} \subset \mathcal{Y}$ proposed by Gittings \emph{et al.}\ \cite{Gittings2019}.  We evenly distribute these attack classes across the test set; let this {\bf set of attack images} be $\mathcal{I}$.

{\bf Metrics.}  We measure the {\bf attack success rate}  as the proportion of  $\mathcal{I}$, containing patches crafted to indicate misclassification as $c \in \mathcal{A}$ result in those image being misclassified as $a$; \emph{i.e.}\ the success  rate of a {\em targeted attack}. We measure the {\bf defence success rate} as the proportion of $\mathcal{I}$ that are correctly classified as  their true class $y$ (despite the APA).  Thus the inverse of the defence success rate, is the {\em untargeted} attack success rate \emph{i.e.}\ where any misclassification occurs due to the APA.  All success rates are expressed as the percentage of the 50k attack image set $\mathcal{I}$ constructed with the APA analysed in that experiment.  All experiments were run for 1000 iterations training and 5 restarts.

\subsection{D-VaN vs Baseline defences}
We first evaluate the performance of our defence (D-VaN) at reducing the effectiveness of adversarial patches synthesised by existing APA attack methods A-ADS \cite{Brown2017} and A-DIP \cite{Gittings2019}.  Both of these methods are white-box attacks, that run backpropagation through the model in order to generate patches to attack it. 

We mount such attacks against our defence, the two baseline defences, and an undefended model as a control. In the case of the baseline methods we use patches that are trained on the undefended network, and then apply them to the defended network, since the defence layers are not usually differentiable. In the case of our model we attack it using patches generated on both the defended and undefended networks; D-VaN(D)/D-VaN(U). This measures transferability of the learned protection against attack from our generator $G$ to the A-ADS and A-DIP attacks. We report both D-VaN(D) and D-VaN(U) because they can each highlight different flaws in a network's defences, and both make sense as real-world attack vectors.
\begin{figure}[t!]
	\centering
	\includegraphics[width=\textwidth]{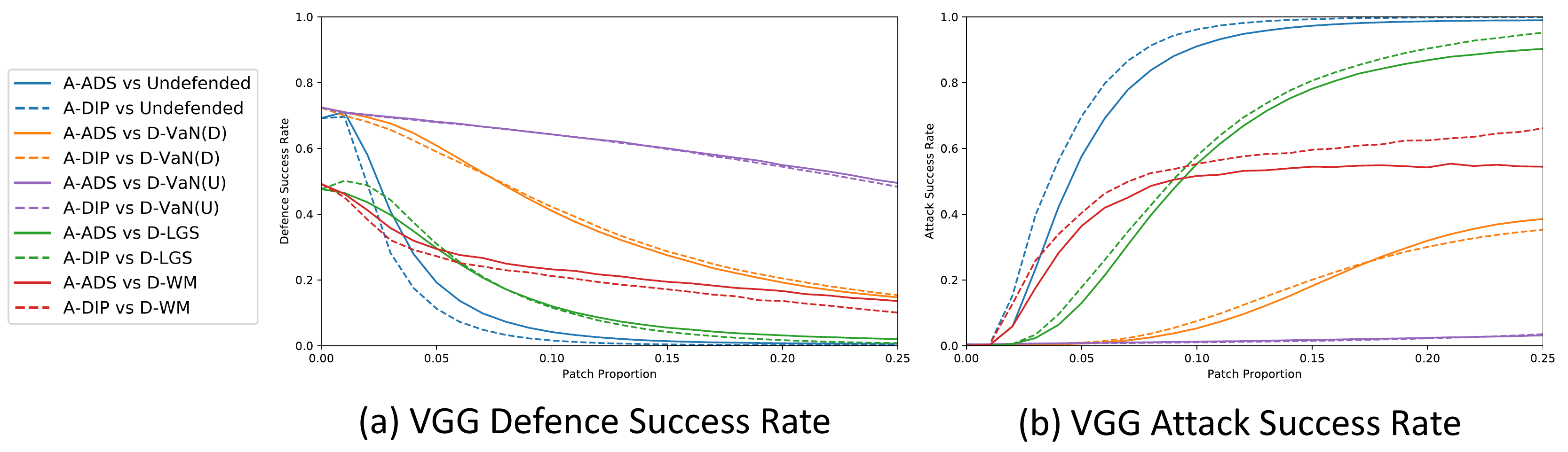}
	\caption{\label{fig:vgg} Success rates of defended VGG-19 networks against APAs for patches covering up to 25\% of the image.}
	\squeezeup
\end{figure}
\begin{figure}[t!]
	\centering
	\includegraphics[width=\textwidth]{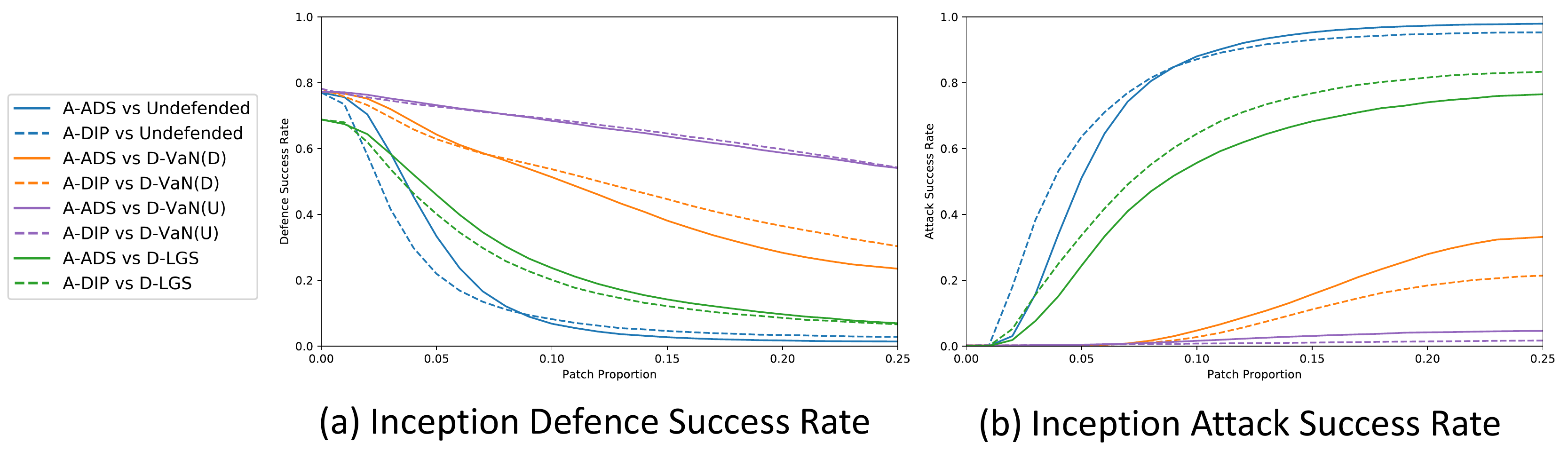}
	\caption{\label{fig:inception} Success rates of defended Inception-v3 networks against APAs for patches covering up to 25\% of the image. We do not include a line for D-WM since the implementation of the defence was incompatible with Inception-v3.}
	\squeezeup
\end{figure}
\begin{figure}[t!]
	\centering
	\includegraphics[width=\textwidth]{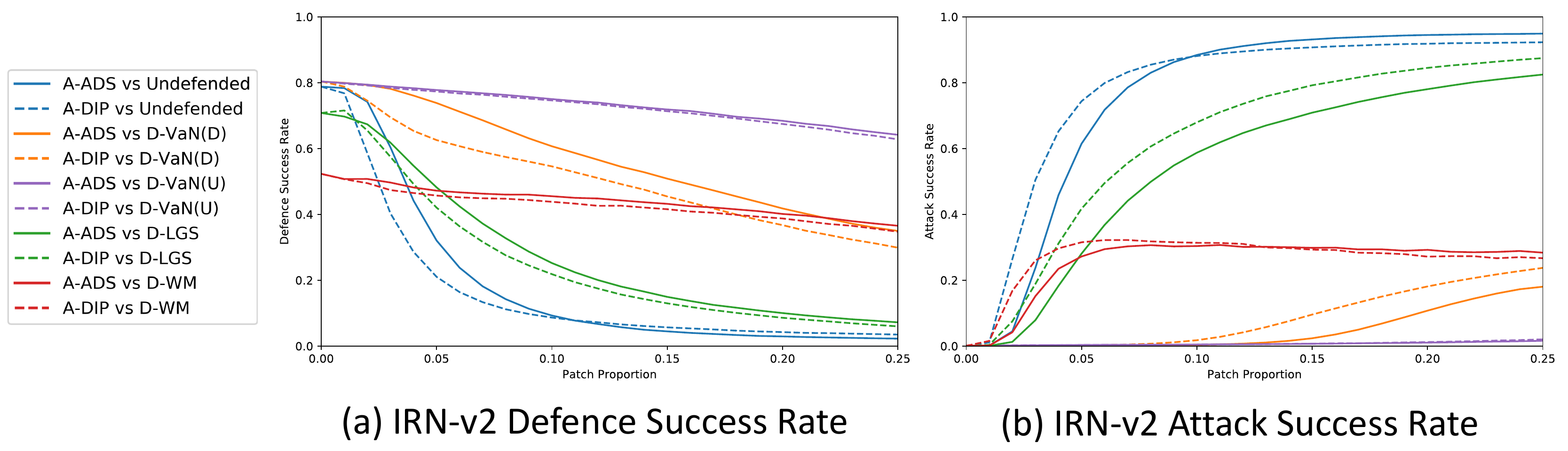}
	\caption{\label{fig:irnv2} Success rates of defended InceptionResNet-v2 networks against APAs for patches covering up to 25\% of the image.}
	\squeezeup
\end{figure}

\begin{figure}[t!]
\captionsetup[subfigure]{font=tiny,labelfont=tiny}
\centering %
\begin{subfigure}[b]{0.19\textwidth}
	\includegraphics[width=\textwidth]{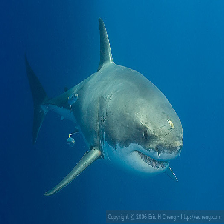}
	\caption{Original image}
\end{subfigure}
\begin{subfigure}[b]{0.19\textwidth}
	\includegraphics[width=\textwidth]{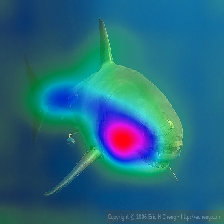}
	\caption{Undef.\ `shark'}
\end{subfigure}
\begin{subfigure}[b]{0.19\textwidth}
	\includegraphics[width=\textwidth]{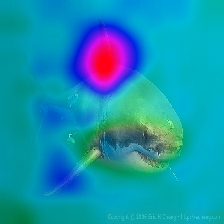}
	\caption{Undef.\ `wallaby'}
\end{subfigure}
\begin{subfigure}[b]{0.19\textwidth}
	\includegraphics[width=\textwidth]{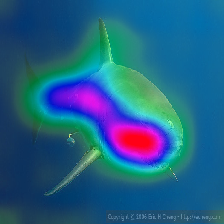}
	\caption{D-VaN `shark'}
\end{subfigure}
\begin{subfigure}[b]{0.19\textwidth}
	\includegraphics[width=\textwidth]{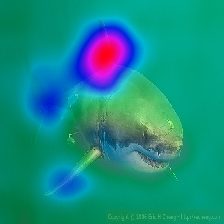}
	\caption{D-VaN `wallaby'}
\end{subfigure}

\medskip 
\begin{subfigure}[b]{0.19\textwidth}
	\includegraphics[width=\textwidth]{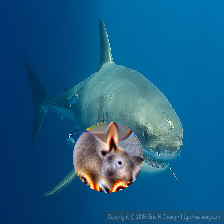}
	\caption{Original image}
\end{subfigure}
\begin{subfigure}[b]{0.19\textwidth}
	\includegraphics[width=\textwidth]{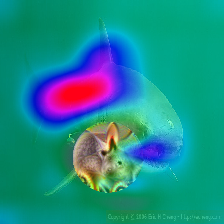}
	\caption{Undef.\ `shark'}
\end{subfigure}
\begin{subfigure}[b]{0.19\textwidth}
	\includegraphics[width=\textwidth]{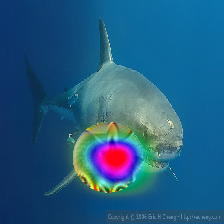}
	\caption{Undef.\ `wallaby'}
\end{subfigure}
\begin{subfigure}[b]{0.19\textwidth}
	\includegraphics[width=\textwidth]{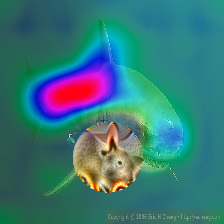}
	\caption{D-VaN `shark'}
\end{subfigure}
\begin{subfigure}[b]{0.19\textwidth}
	\includegraphics[width=\textwidth]{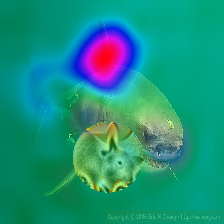}
	\caption{D-VaN `wallaby'}
\end{subfigure}
\caption{\label{fig:gradcam} Grad-CAM \cite{gradcam} visualisations for our VaN defended VGG network vs the undefended model. The original image (a) is classified (correctly) as a great white shark by both the undefended and defended (D-VaN) models, whereas the patch image is misclassified as a shark by the undefended model, but classified correctly by the defended (D-VaN) model.}
	\squeezeup
\end{figure}

We consider patches of a variety of sizes up to $25\%$ of the total image area.  Patches are placed randomly, anywhere in the image, and with a random rotation of up to \(\sfrac{\pi}{8}\) for all experiments.

In Table~\ref{tbl:control} we report the accuracy of our model and all the baseline models on images with no adversarial attack, for \(\mathcal{Y}\) and \(\mathcal{A}\). The two baseline defences substantially reduce the accuracy of the model on the unattacked images, which is very significant for most applications since adversarial examples are relatively rare, \emph{i.e.}\ clean images represent the overwhelming majority of samples that will be encountered in the real world. Our defended network maintains the accuracy of the undefended classifier on this set for all 3 classifiers we tested. We also note that no defence method significantly reduces model sensitivity for $\mathcal{A}$ given clean images, which could cheat the trial by failing to ever identify images as these adversarial test classes. 

\subsection{Network Architecture and Patch Size}

Table~\ref{tbl:defence_success} reports the improved resilience of models under our defence, showing significantly higher defence success rates for VGG, Inception and IRN-v2 architectures at $41.0\%$, $51.3\%$, and $60.7\%$ and $14.7\%$, $23.5\%$, and $35.0\%$ respectively for smaller and larger patches in the case of D-VaN(D). For smaller patches these rates are at least $30\%$ higher than the closest baseline defence method, and for larger patches they are comparable. If we consider instead D-VaN(U), then for smaller patches the accuracy is reduced by only at most $25\%$ from the original, and for larger patches it is still greater than $60\%$ of its original value. 

Table~\ref{tbl:attack_success} shows the reduced vulnerability of our defended models, for all 3 architectures. Again the reduction is most evident for smaller patches, where our defended classifier is fooled less than $10\%$ as often as our closest competitor. The performance at for larger patch sizes is closer, but we still outperform baselines. In the case of D-VaN(U), our attack success rate is reduced to less than $5\%$ for all networks, even for the largest patches.

 Figs.\ \ref{fig:vgg}, \ref{fig:inception} and \ref{fig:irnv2} show the dependence of attack and defence success rates on size, for our defence method as well as the baseline methods. Our method is an effective defence for all three architectures we are testing, and at all scales of patch. The performance of our method degrades as the size of the patch increases, which is expected since the patch covers up to 25\% of the image, possibly occluding some salient object detail.
 

\subsection{Attention under attack}

Fig.\ \ref{fig:gradcam} uses Grad-CAM \cite{gradcam} to localise CNN attention for a particular class, for both our D-VaN defended model and the undefended model. Here the model is being attacked via A-ADS with target class of `wallaby' whereas the true class of the image is `great white shark'.  Note that all plots are normalised; blue/purple relatively high attention, green/blue relative low.  For images flooded with green/blue, there was low response for that class (c,e,g,j). 

On the original image (a) with no patch, our model (d) and the undefended model (b) perform similarly. Both decide on the most likely class as shark, and  both identify the region containing the shark as being of high importance. For this unattacked image, the response for the counterfactual class `wallaby' is naturally low and both (c,e) pick a somewhat arbitrary area in the image that was of low importance to the correct decision (shark). 

When the adversarial patch A-ADS targeting the counterfactual class is introduced (lower row), the undefended model  identifies that patch region as very high salience for the wallaby class (h) and decides on wallaby, whereas our D-VaN defended model does not change its decision from shark, and does not attend to the patch (i).  Forcing Grad-CAM to explain shark for the undefended model  (which was not the decision outcome, so produces low attention) the original model picks out the area of the shark unoccluded by the patch (g) as does our defended model  (i).  In our case the model can correctly identify the shark, but in the original case it cannot since its attention was attracted by the wallaby patch.  For completeness we show the defended model does not localise wallaby even when forced to explain wallaby in the attacked image (j).

\begin{figure}[t!]
	\centering
	\includegraphics[width=\textwidth]{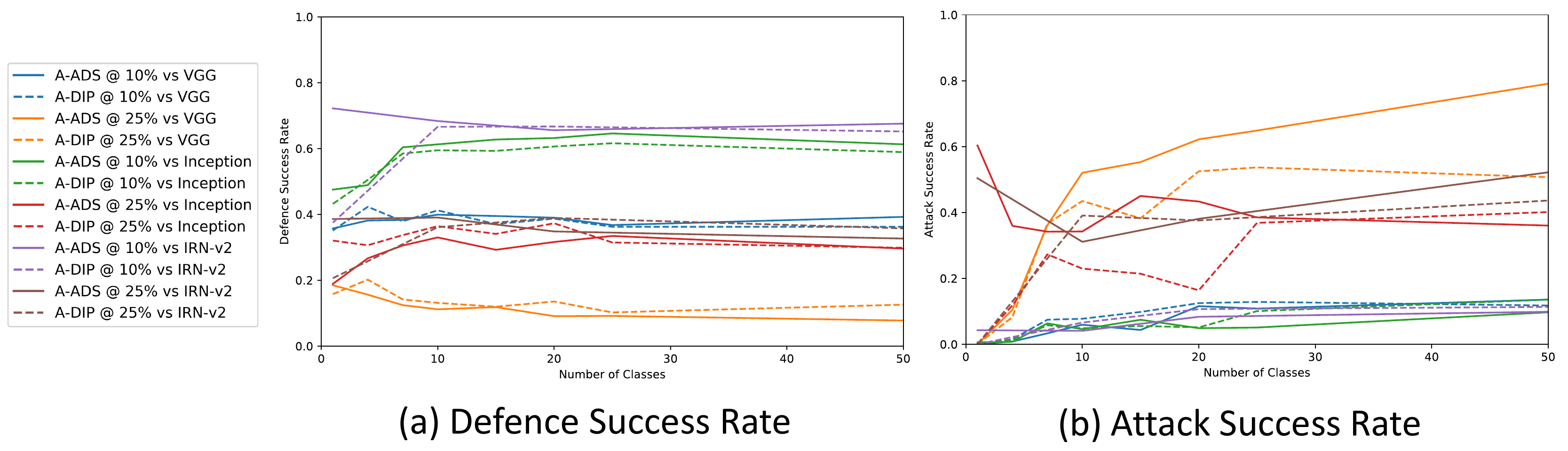}
	\caption{\label{fig:num_classes} Success rate of defended networks as we vary the number of classes of APA that our conditional generator produces. }
	\squeezeup
\end{figure}
\subsection{Class Generalization}
\label{sec:classexp}
In Fig.\ \ref{fig:num_classes} we examine the effect of changing the number of classes which our conditional generator produces. We train each of the three network architectures to defend against between 1 and 50 classes of adversarial patch, and we evaluate their performance against both A-ADS and A-DIP attacks with patches taking up 10\% or 25\% of the image. We find that the defence success rate is consistent as the number of classes changes for each network and for each patch size, showing that our method does not break down as the number of classes is increased. For the attack success rate we note that for the most part it increases slightly as the number of classes increases. The exception is large A-ADS patches on Inception-v3 and InceptionResNet-v2 architectures, for which our model loses performance when targeting a very small number of classes. This suggests value in the attack class diversity availabile during training due to our conditional patch generator $G$.

\subsection{Timing Information}
Table~\ref{tbl:inference_timing} compares the time taken for inference  using our method and baselines. An inference pass on the defended model takes the same time as on the undefended model; the architecture is unchanged. However our defence does take 2-3 hours of training to `vaccinate' the model.  This process only needs to be run once, as does training the model {\em a priori}. The baseline APA defences run as a pre-process at inference time, and so take longer (and also degrade accuracy; Table~\ref{tbl:control}).  All runs used an NVIDIA GeForce GTX 1080 Ti GPU.

\begin{table}[t!]
\centering
\small
\caption{Inference time (seconds) for an undefended VGG model trained on ImageNet, and that model with our defence or  baseline defences applied.}
\label{tbl:inference_timing}
\begin{tabular}{l|ccc}
\toprule 
Method & VGG & Inception & IRN-v2 \\
    \midrule
Undefended & \textbf{0.10} & \textbf{0.08} & \textbf{0.16} \\
D-VaN/Ours & \textbf{0.10} & \textbf{0.08} & \textbf{0.16} \\
D-WM & 1.35 & - & 1.08 \\
D-LGS & 0.32 & 0.40 & 0.45 \\
\bottomrule
\end{tabular}
\end{table}

\subsection{Physical Experiment}
To test the effectiveness of our defence against attacks in the physical world, where the appearance of the patch could differ from its digital form, we generated a patch to attack a VGG network targeting ImageNet class 964 ``potpie'' using A-ADS. We placed this patch on or around objects of 47 different ImageNet classes found in the physical world, for a total of 126 photographs of a patch. The photos were taken on a Google Pixel 2 smartphone. The undefended classifier returned adversarial vs. correct class 84 vs 9 times (attack success rate 90.3\%), whereas the Vax-A-Net defended classifier returned similarly 5 vs 71 (attack success rate 6.6\%).

\section{Conclusion}
We proposed Vax-a-Net; a method to `immunise' (defend) CNN classifiers against adversarial patch attacks without degrading the performance of the model on clean data and without slowing down the inference time. In the process of achieving this we produced a conditional generator for adversarial patches, and then we used an adversarial training methodology to update the generator during training rather than having to synthesise patches from scratch at each iteration. We showed experimentally that our method performs better than the baseline defences in both a targeted and untargeted sense, and across three different popular network architectures. Furthermore we showed that our network is resilient to patches produced by two different attacks, and to patches that are produced either on our defended network or on the original undefended network, which demonstrates that our defence taught the network real robustness to these patches, and not simply to hide its gradient or to ignore a group of specific patches. Future work could look into extending these methodologies to defend networks for different tasks, such as patch attacks on object detectors.

\section*{Acknowledgment}
The first author was supported by an EPSRC Industrial Case Award with Thales, United Kingdom.  

\bibliographystyle{splncs04}
\bibliography{references}
\end{document}